%% file: bare_jrnl_compsoc.tex
\documentclass[10pt,journal,twocolumn]{IEEEtran}
\ifCLASSOPTIONcompsoc
  \usepackage[nocompress]{cite}
\else
  \usepackage{cite}
\fi

\ifCLASSINFOpdf
\else
\fi

\usepackage{times}
\usepackage{epsfig}
\usepackage{graphicx}
\usepackage{amsmath}
\usepackage{amssymb}
\usepackage{eucal,bibspacing}
\usepackage{cs}
\usepackage{mathsymb}
\usepackage{textcomp}
\usepackage{verbatim}
\usepackage{url}
\usepackage{multirow}
\usepackage[super]{nth}
\usepackage{colortbl}
\usepackage{color}
\usepackage{pbox}
\usepackage{xcolor}
\usepackage{booktabs,makecell}
\usepackage{mathtools}

\usepackage{wrapfig}
\usepackage{caption}
\usepackage[linesnumbered,algoruled,boxed,lined]{algorithm2e}

\def\bx{{\boldsymbol x}}

\let\savedalgorithm\algorithm
\let\savedendalgorithm\endalgorithm

\usepackage{algorithm}

\renewcommand{\etal}{\textit{et al. }}

\newcommand{\ie}{\textit{i.e., }}

\begin{document}

\title{3D PersonVLAD: Learning Deep Global Representations for Video-based Person Re-identification}

\author{Lin Wu, Yang Wang, Ling Shao, \textit{Senior Member, IEEE} and Meng Wang, \textit{Senior Member, IEEE}
\IEEEcompsocitemizethanks{\IEEEcompsocthanksitem Lin Wu is with The University of Queensland, St Lucia 4072, Australia. Email: lin.wu@uq.edu.au.

Yang Wang* (Corresponding author) is with School of Computer Science and Information Engineering, Hefei University of Technology, Hefei 230000, China and Dalian University of Technology, China.
Email: yang.wang@dlut.edu.cn.

Ling Shao is with Inception Institute of Artificial Intelligence (IIAI), Abu Dhabi, UAE. Email: ling.shao@ieee.org.

Meng Wang is with the School of Computer Science and Information Engineering, Hefei University of Technology, Hefei 230000, China. Email: eric.mengwang@gmail.com. \protect\\

}
}

\IEEEtitleabstractindextext{%
\begin{abstract}
In this paper, we introduce a global video representation to video-based person re-identification (re-ID) that aggregates local 3D features across the entire video extent. Most of the existing methods rely on 2D convolutional networks (ConvNets) to extract frame-wise deep features which are pooled temporally to generate the video-level representations. However, 2D ConvNets lose temporal input information immediately after the convolution, and a separate temporal pooling is limited in capturing human motion in shorter sequences. To this end, we present a \textit{global} video representation (3D PersonVLAD), complementary to 3D ConvNets as a novel layer to capture the appearance and motion dynamics in full-length videos. However, encoding each video frame in its entirety and computing an aggregate global representation across all frames is tremendously challenging due to occlusions and misalignments. To resolve this, our proposed network is further augmented with 3D part alignment module to learn local features through soft-attention module. These attended features are statistically aggregated to yield identity-discriminative representations. Our global 3D features are demonstrated to achieve state-of-the-art results on three benchmark datasets: MARS \cite{MARS}, iLIDS-VID \cite{VideoRanking}, and PRID 2011 \cite{PRID2011}.
\end{abstract}

}

\maketitle

\IEEEdisplaynontitleabstractindextext

\IEEEpeerreviewmaketitle

\section{Introduction}\label{sec:intro}

\IEEEPARstart{P}{erson} re-identification (re-ID) refers to matching pedestrians in the context of non-overlapping camera views. It is attracting substantial attention in computer vision due to its wide range of potential applications, such as human behavior analysis and security in public places. In general, the existing person re-ID methods can be categorized into image-based approaches and video-based alternatives. Many approaches based on deep neural networks belong to the image-based stream \cite{GatedCNN,Multi-channel-part,E-Metric,S-LSTM,DGDropout,SpindleNet,SimilaritySpatial,Correspondence,LOMOMetric,Part-Aligned,
SI-CI,PIE-reid,Deep-Embed,SSM,GOG,LDA-Re-ID,Semi-Coupled-DL,PR18YangLIN,TIP17Yang,High-order-Re-ID,Multi-level-Re-ID,Good-Practice-Re-ID} while relatively fewer works deal with videos \cite{Video-person-ijcai16,Top-push,RCNRe-id,ACMMM15Yang,RFA-net,VideoRanking,VideoPerson,Video-end-end,Joint-spatial-temporal}. In practice, the video input provides a more natural solution to person recognition because pedestrian videos can be easily captured in a surveillance system. Moreover, videos contain richer information than images, and beneficial for identifying a person under complex conditions.

\begin{figure}[t]
\centering
\includegraphics[width=8.5cm,height=4.5cm]{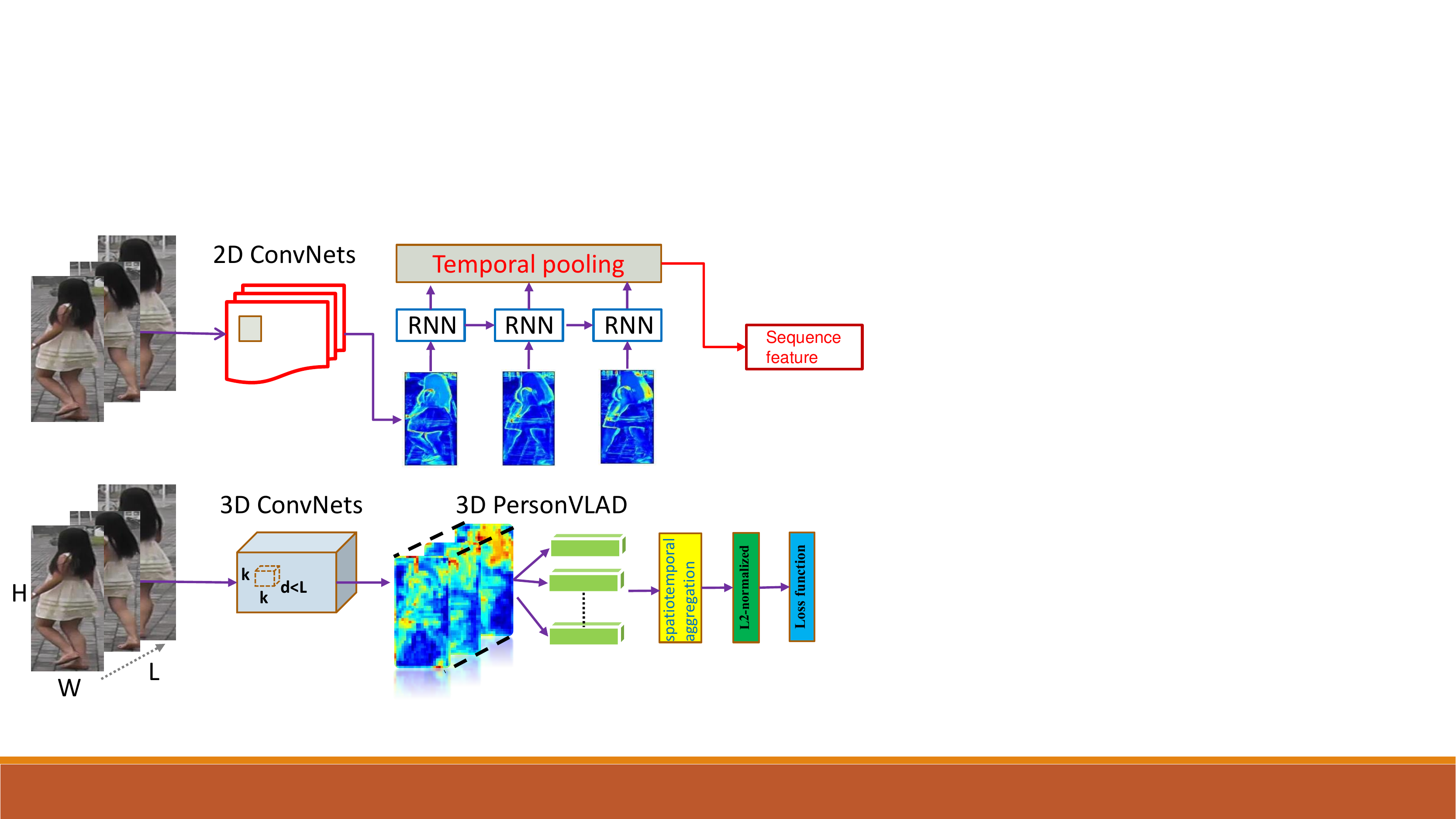}
\caption{Top: Existing methods based on 2D ConvNets extract deep features frame-wise which are pooled temporally (e.g. through RNNs) to produce video-level features. However, temporal priors are completely lost right after each convolution and 2D ConvNets are incapable of modeling a long video sequence. Bottom: The proposed method learns a global video-level representation across the entire spatiotemporal extent with efficiency.}\label{fig:3d-features}
\end{figure}

Recent video-based re-ID methods focus on learning deep features by considering both spatial and temporal information \cite{Diversity-Video-Re-ID}. Some spatial networks are based on one-stream ConvNets to perform appearance modeling from individual frames, and appearance features are pooled temporally to yield the sequence-level representations \cite{Video-end-end,RCNRe-id,RFA-net}. However, frame-based deep features are not suitable for videos due to the lack of motion modeling, and critically 2D ConvNets lose temporal priors of the input immediately after the convolutions. Some current studies demonstrate that two-stream ConvNets \cite{Two-stream-re-id,Joint-spatial-temporal,ASTPN} outperform one-stream ones by decomposing the videos into motion and appearance streams. Then separate CNNs for each stream are trained and the outputs are fused in the end. However, current CNN methods for video-based re-ID often extend CNN architectures designed for static images and learn the representations for short video intervals. Yet, pedestrian videos often contain complex spatiotemporal characteristics with specific spatial as well as long-term temporal structure. The problem even becomes exaggerated when persons have large intra-class variability caused by the changes of viewpoints, illuminations and human poses. Breaking this structure into short clips, and aggregating video-level information by the simple average of clip scores or recurrence scheme such as LSTMs \cite{RFA-net} is likely to be suboptimal.

\subsection{Our Approach}
To solve the problem above,  in this paper, we propose to learn the long-term, global representations for video-based person re-ID. We consider space-time ConvNets \cite{3DConvNets} to perform 3D convolutions/pooling, and study the architecture that aggregates mid-level convolutional descriptors across different portions of the imaged identity across the entire temporal span. It is known that 3D ConvNets are capable of modeling the temporal prior much better because convolution and pooling are performed spatiotemporally as opposed to 2D ConvNets with only spatial modeling. The core to our network goes to the proposed \textbf{3D PersonVLAD} aggregation layer, that is inspired by the ``Vector of Locally Aggregated Descriptors" \cite{All-VLAD,VLAD}. This 3D pooling operation captures information about the statistics of local convolutional features aggregated over the full-length temporal extent. The resulting aggregated representation is regarded as the \textit{global} descriptor for the person video. However, one potential limitation of using global representations lies in the absence of many explicit mechanism to tackle the misalignment inherent to human pose changes and imperfect detectors.

In view of attention models \cite{ShowAttendTell} and current reflection on human body region based representation learning in static-image person re-ID \cite{Multi-channel-part,S-LSTM,SpindleNet,Part-Aligned,Context-body}, we propose to learn 3D spatiotemporal representations with body part alignment. Following this way, the 3D body filters with attention are able to select the most informative regions both spatially and temporally, and propagate them into the aggregation layer to form a global representation. The attention module is complementary to 3D convolutions and helps with modeling long-range discriminations effectively.\\
\textbf{Remark. }We remark some major contributions of our model upon the existing attention models in person re-ID \cite{Diversity-Video-Re-ID,Joint-spatial-temporal,ASTPN,Hydraplus}. First, current deep attention methods commonly employ different attention mechanisms to spatial and temporal priors. For instance, in \cite{Diversity-Video-Re-ID} a penalization term is used to regularize spatial attention whilst temporal attention is implemented by assigning weights to different salient regions. In contrast, our method has integrated spatiotemporal attention whose parameters are learned over the course of end-to-end training of the entire network. Second, in assembling local features into global, these attention models are developed without consideration of the compatibility of local and global features. To this end, we design the architecture with the VLAD pooling that statistically captures the information of local descriptors over the temporal extent. Thus, the network is forced to produce global presentations discriminative yet robust to visual variations, while simultaneously learns to focus on the relevant parts of the identities.

\subsection{Contributions}

We summarize the contributions of this work as follows.

\begin{itemize}
\item We present 3D PersonVLAD aggregation for video-based person re-ID that attends human body part appearance and motion simultaneously while encoding these interactions into its global representation of a full-length video.

\item We demonstrate the advantages of integrated spatiotemporal attention and the importance of VLAD aggregation for learning accurate, generic, efficient video representations for person re-ID.

\item The proposed architecture significantly outperforms off-the-shelf CNN descriptors and two-stream image/motion representations on three challenging benchmarks, and improves over the current best methods on benchmarks. We advance the state-of-the-art results from 71.7\% to \textbf{80.8}\% on MARS \cite{MARS}, 62.0\% to \textbf{69.4}\% on iLIDS-VID \cite{VideoRanking}, and 79.4\% to \textbf{87.6}\% on PRID 2011 \cite{PRID2011}.
\end{itemize}

The rest of this paper is structured as follows. In Section \ref{sec:related}, we briefly review related works. Section \ref{sec:approach} presents the architecture of the proposed method. Extensive experimental results on three public data sets are given in Section \ref{sec:exp} and Section \ref{sec:con} concludes this paper.

\input{related.tex}

\input{approach.tex}

\input{experiment.tex}

\section{Conclusions}\label{sec:con}

In this paper, we present an end-to-end trainable deep neural network to produce global video-level features over the entire video span for video-based person re-ID. The basic idea is to build up a deep 3D convolution architecture with an amenable layer, namely 3D PersonVLAD aggregation layer, to encode spatiotemporal signals into a global compact video descriptor. To make the learned video features discriminative against various misalignment, a 3D part alignment module based on attention models is introduced into the feature learning stage to localize distinct regions from which local features are aggregated to yield the global representations highly robust to pose changes and various human spatial distributions. With the global space-time convolutions over full-length video extent, we obtain the state-of-the-art performance over three video-based person re-ID datasets. We also demonstrate the generalization and scalability of the proposed model.


\ifCLASSOPTIONcaptionsoff
  \newpage
\fi



\bibliographystyle{IEEEtran}\small
\bibliography{allbib}

\begin{IEEEbiography}{Lin Wu} was awarded a PhD from The University of New South Wales, Sydney, Australia in 2014. She has published 40 academic papers, such as CVPR, ACM Multimedia, IJCAI, ACM SIGIR, IEEE-TIP, IEEE-TNNLS, IEEE-TCYB, and Pattern Recognition. She also regularly served as the program committee member for numerous international conferences and invited journal reviewer for IEEE TIP, IEEE TNNLS, IEEE TCSVT, IEEE TMM and Pattern Recognition.
\end{IEEEbiography}

\begin{IEEEbiography}{Yang Wang} earned the Ph.D. degree from
The University of New South Wales, Kensington,
Australia, in 2015. He is currently a Professor at Hefei University of Technology, China,
Before that, he has been an Associate Professor at Dalian University of Technology, China.
He has published 50 research
papers, most of which have appeared at the competitive
venues, including IEEE TIP, IEEE TNNLS,
IEEE TCYB, IEEE TKDE, IEEE TMM, Pattern
Recognition, Neural Networks, ACM Multimedia,
ACM SIGIR, IJCAI, IEEE ICDM, ACM CIKM, and
VLDB Journal. He served as the Invited Journal
Reviewer for more than 15 leading journals, such
as IEEE TPAMI, IEEE TIP, IEEE TNNLS, IEEE
TKDE, Machine Learning (Springer), and IEEE TMM.
\end{IEEEbiography}

\begin{IEEEbiography}{Ling Shao} is the CEO and the Chief Scientist
of the Inception Institute of Artificial Intelligence,
Abu Dhabi, United Arab Emirates. His research
interests include computer vision, machine learning,
and medical imaging. He is a fellow of the IAPR,
the IET, and the BCS. He is an Associate Editor
of the IEEE TRANSACTIONS ON IMAGE PROCESSING,
the IEEE TRANSACTIONS ON NEURAL
NETWORKS AND LEARNING SYSTEMS, and several
other journals.
\end{IEEEbiography}

\begin{IEEEbiography}{Meng Wang} is a professor at the Hefei University of Technology, China. He received his B.E. degree and Ph.D. degree in the Special Class for the Gifted Young and the Department of Electronic Engineering and Information Science from the University of Science and Technology of China (USTC), Hefei, China, in 2003 and 2008, respectively. His current research interests include multimedia content analysis, computer vision, and pattern recognition. He has authored more than 200 book chapters, journal and conference papers in these areas. He is the recipient of the ACM SIGMM Rising Star Award 2014. He is an associate editor of IEEE Transactions on Knowledge and Data Engineering (IEEE TKDE), IEEE Transactions on Circuits and Systems for Video Technology (IEEE TCSVT), IEEE Transactions on Neural Networks and Learning Systems (IEEE TNNLS), and IEEE Transactions on Multimedia (IEEE TMM).
\end{IEEEbiography}

\end{document}

%% file: related.tex
\section{Related Work}\label{sec:related}

\subsection{Video-based Person Re-identification}
An image sequence can be viewed as a 3-dim space-time volume and space-time features can be extracted based on space-time interest points \cite{3DSIFT}. In person re-ID, some works focus on motion features to describe the appearance variations of pedestrians. For instance, Wang \etal \cite{VideoRanking} employ the HOG3D \cite{HOG3D} descriptor with dense sampling after identifying walking periodicity. However, a pre-processing is needed to select discriminative video fragments from which HOG3D features can be extracted. Another work from Liu \etal \cite{VideoPerson} is to extract 3D low-level features that encode both spatially and temporal aligned appearance of a pedestrian. They also need to manually align video fragments by using regulated flow energy profile and spatial alignment is achieved by partitioning human body into six rectangles to describe different body parts.

More recently, deep neural networks are demonstrated to be effective for image based person re-ID and have achieved notable improved performance \cite{GatedCNN,Multi-channel-part,S-LSTM,DGDropout,TMM18Lin,SpindleNet,Part-Aligned,SI-CI,Deep-Embed,LDA-Re-ID}. In contrast, video-based re-ID is paid less attention. McLaughlin \etal \cite{RCNRe-id} extract CNN features for each frame, and put through a RNN to impose their temporal dependency. In \cite{Video-end-end}, an end-to-end approach is proposed to simultaneously learn deep features and their similarity metric for a pair of videos as input. Also Yan \etal propose a recurrent feature aggregation network (RFA-Net) \cite{RFA-net} which shares the similar idea of using RNN to aggregate frame-wise features into a sequence level representation. Another effective approach \cite{Joint-spatial-temporal} is to use two RNNs where one is to exploit the temporal dimension and the other is to calculate the spatially context similarity of two corresponding locations in a pair of video sequences. Similar to \cite{Joint-spatial-temporal}, an attentive spatial-temporal pooling is presented \cite{ASTPN} to employ spatial pooling to select regions while temporal pooling is used to select informative frames.

To go beyond individual image-level appearance information and exploit the temporal information, Simonyan \etal \cite{Two-stream-fusion} propose the two-stream architecture, cohorts of spatial and temporal ConvNets. The input to the spatial and temporal networks are RGB frames and stacks of multiple-frame dense optical flow fields, respectively. Wang \etal \cite{Temp-segment} further improve the two-stream architecture by enforcing consensus over predictions in individual frames. These networks are still limited in their capacity to capture temporal information, because they operate on a fixed number of regularly spaced, single frames from the entire video. The ActionVLAD \cite{ActionVLAD} extends the NetVLAD aggregation layer \cite{NetVLAD} to videos and shows that aggregating the last layers of convolutional descriptors regarding spatial and temporal networks performs better. Inspired by these two-stream architectures, a two stream Siamese convolutional neural network architecture \cite{Two-stream-re-id} is proposed to use two CNNs to process spatial and temporal information separately, and the outputs of two streams are fused together using a weighted cost function. However, these deep learning based methods commonly employ 2D ConvNets to extract feature activations for each frame which are leveraged into RNNs to ensure the temporal dependency among frames. In this way, temporal priors are completely collapsed right after convolutions. In contrast to these methods, we develop an end-to-end trainable video architecture that combines the recent advances in 3D ConvNets with a trainable spatiotemporal part alignment based VLAD aggregation layer. This is to our best knowledge, has not be done before. In addition, we compare favourably the performance of our method with the above methods in Section \ref{ssec:compare-ConvNets}.

\subsection{Attention Models in Person Re-identification}

The attention mechanism has become a popular practice in visual-textual relationship \cite{ShowAttendTell} and person recognition \cite{Diversity-Video-Re-ID,Joint-spatial-temporal,TCYB18,ASTPN,Hydraplus}.
Li \etal \cite{Diversity-Video-Re-ID} propose a spatiotemporal attention scheme to video-based re-ID that uses multiple spatial attention models to localize discriminative image regions, and pool these extracted local features across time using temporal attention. However, they assemble an effective representation regarding a person extracted from multiple spatial attention models. In this way, multiple weights for temporal pooling are required to be learned. Zhou \etal \cite{Joint-spatial-temporal} combine the spatial and temporal information by building an end-to-end attention network that assigns importance scores to input frames, in accordance with the hidden states of an RNN. The final feature is produced by average pooling of the RNN's outputs. However, in the training of their model, the weights at each time step tend to have the same values.

\subsection{3D Convolutional Networks}

Quite a few are concerned with effective representations using local spatiotemporal features. For example, Laptev \etal \cite{3D-STIP} proposed spatiotemporal interest points (STIPs) by extending Harris corner detectors to 3D. SIFT and HOG are also extended to be SIFT-3D \cite{3DSIFT} and HOG3D \cite{HOG3D} for action recognition. Improved Dense Trajectories \cite{iDT} are the state-of-the-art hand-crafted features. Nonetheless, it starts with densely sampled features points in video fragments and uses optical flows to track them. This is computationally intensive and intractable on large-scale data sets. With the availability of large amount of training data and powerful GPUs, convolutional neural networks (ConvNets) have achieved breakthroughs in visual recognition \cite{AlexNet}, and these deep networks are used for image feature learning. Deep learning has also been applied to video feature learning in supervised \cite{Karpathy} or unsupervised setting \cite{VideoLSTM,DBLP:journals/tip/WangLWZZH15}. Recently, 3D ConvNets are proposed for human action recognition \cite{3DConvNets,LTC} and person re-ID \cite{Kato-ICIAR-2018}. However, there is no principled approach to exploit 3D deep nets for video-based person re-ID to learn a global discriminative representation across the entire video sequence.

Our work is also related to feature aggregation such as VLAD \cite{ActionVLAD,NetVLAD} and Fisher vectors (FV) \cite{LDA-Re-ID}. In recent, these aggregation techniques have been extended into end-to-end training within a CNN for representing images in different scenarios, such as instance-level retrieval \cite{Deep-IR} and place recognition \cite{NetVLAD}. We build on this work and develop it to be an end-to-end trainable video representation for person re-ID by feature aggregation over human body part alignment over the entire video span.

%% file: approach.tex
\begin{figure*}[hbt]
\centering
\includegraphics[width=14cm,height=3cm]{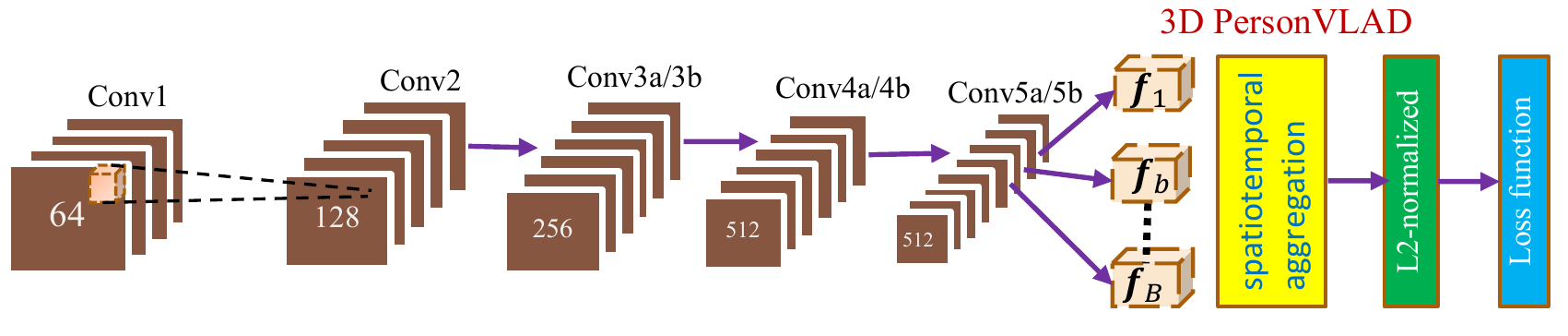}
\caption{The proposed 3D PersonVLAD architecture for video-based person re-ID. The network has 8 convolutions, 5 max-pooling, 3D body part alignment (consists of $B$ part map detectors), followed by the aggregation layer and a loss function layer. Spatiotemporal convolutions with $3\times 3 \times 3$ are applied into the convolutions. Max-pooling and ReLU are applied between all convolutional layers.}\label{fig:network}
\end{figure*}

\section{Learning Global Discriminative Spatiotemporal Features with 3D Body Part Aggregation}\label{sec:approach}

In this section, we present the network architecture to seek a video-level global representation that is end-to-end trainable. We introduce an architecture outlined in Fig.\ref{fig:network}. In detail, we sample 3D convolutional features from the entire video, and aggregate body features with alignment using a vocabulary of ``person words" into a single video-level fixed length vector. This representation is then put through a classifier that outputs the final classification scores on each identity. The parameters of the aggregation layer, i.e., the set of ``person words" are learnt jointly with the feature extractors in a discriminative manner for the task of person re-ID. In what follows, we first describe the basic operations of 3D convolution and pooling for high-level feature extraction (Section \ref{ssec:3D-ConvNets}), then we discuss the proposed 3D body part alignment followed by the aggregation layer (Section \ref{ssec:aggregation}). Finally, the training details are given in Section \ref{ssec:training}.

\subsection{3D Convolution and Pooling}\label{ssec:3D-ConvNets}

It has been shown that 3D ConvNets \cite{3DConvNets}, namely C3D, is well-suited for spatiotemporal feature learning owing to the encapsulated 3D convolution and 3D pooling operations. We follow the network architecture of 3D ConvNets \cite{3DConvNets} which have empirically identified a good architecture for video analysis. A video clip is referred to have a size of $C\times L\times H\times W$ where $C$ is the number of channels, $L$ is the length of frames, $H$ and $W$ are the height and width of the frame, respectively. The 3D convolution and pooling kernel size is denoted by $d\times k\times k$, where $d$ is the kernel temporal depth and $k$ is the kernel spatial size. The networks are set up to take video clips as inputs an predict the class labels which belong to different identities. The 3D convolution and pooling operations are identical to 3D ConvNets \cite{3DConvNets}. The network consists of 5 convolution layers and 5 pooling layers (each convolution layer is immediately followed by a pooling layer), 3D body part alignment layer, the spatiotemporal aggregation layer and a loss function layer to predict the person identity labels. Fig.\ref{fig:network} illustrates the network architecture of our approach. The number of filters for 5 convolution layers are set to be 64, 128, 256, 256, and 256, respectively. All convolution kernels have the size of $d$. All pooling layers are max pooling with kernel size $2\times 2\times 2$ (except for the first pooling layer with kernel size $1\times 2\times 2$).

\begin{figure}[t]
\centering
\includegraphics[width=8cm,height=3cm]{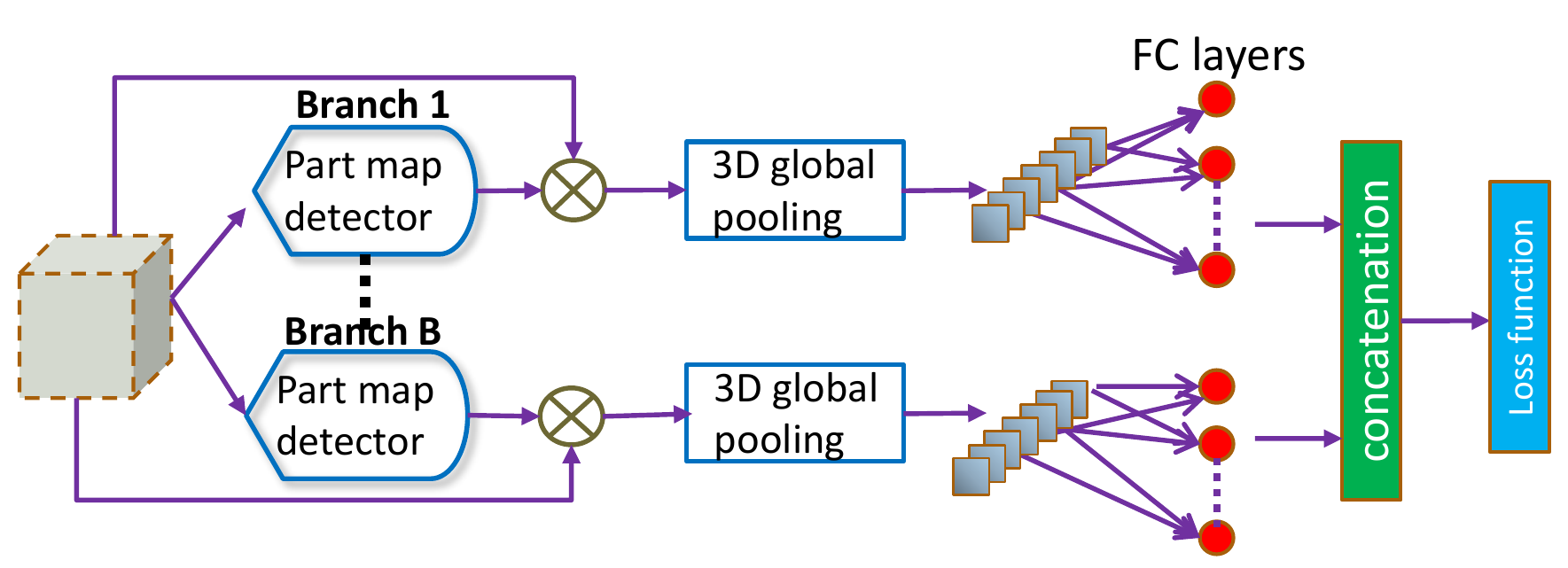}
\caption{The 3D part alignment model consists of $B$ branches, each of which estimates a part attention map to re-weight the feature cubic from 3D convolutions. A global pooling is performed to train the network with localization on the attentive part. }\label{fig:3d-parts}
\end{figure}

\subsection{Trainable Spatiotemporal Aggregation on 3D Body Parts}\label{ssec:aggregation}

To learn part-aligned representations from video clips, we design a 3D body part net which detects, localizes part maps from 3D convolutions, and outputs the part features extracted over the parts. This is inspired by attention models \cite{ShowAttendTell} that are capable of learning salient features from a lot of clutter in an image. The 3D body part net is shown in Fig. \ref{fig:3d-parts}. It contains $B$ branches which correspond to $B$ part detectors. Each branch receives the feature cubic from pool5, detects a discriminative region, and extracts the feature over the detected region as the output, followed by their respective fully-connected layers.

Let a 4-dim tensor $\mathbf{F}$ represent the feature cubic computed from the pool5 and thus $\mathbf{f}(c,l,x,y)$ reflects the $c$-th response to the location $(x,y)$ of the $l$-th frame. The part map detector estimates a 3-dimensional cubic $\mathbf{M}_b$, where $\mathbf{m}_b(l,x,y)$ indicates the degree that the location $(x,y)$of frame $l$ appears in the $b$-th region, from the video feature cubic $\mathbf{F}$. Thus, we have:
\begin{equation}
\mathbf{M}_b= \mathcal{C}_{Detector_b} (\mathbf{F}),
\end{equation}
where $\mathcal{C}_{Detector_b} (\cdot)$ is a region map detector implemented as a 3D convolutional network. Specifically, the part estimator $\mathcal{C}_{Detector_b} (\cdot)$ is implemented as a $1\times 1\times 1$ convolution layer followed by a nonlinear sigmoid layer. The number of part detector $B$ is determined by cross-validation and empirically studied in Section \ref{ssec:analysis}. To detect a region, it is suggested to generating a positive weight for each location, which can be interpreted as the relative importance to give the location where the region would appear \cite{ShowAttendTell}. Thus, the part feature cubic $\mathbf{F}_b$ for the specific $b$-th region can be computed through a weighting scheme below
\begin{equation}
\mathbf{f}_b(c,l,x,y)=\mathbf{f}(c,l,x,y)\otimes \mathbf{m}_b(l,x,y).
\end{equation}
where $\otimes$ indicates element-wise multiplication.

To aggregate these detected body parts into a compact representation over the entire video extent, we consider each feature cubic $\mathbf{f}_b \in \mathbb{R}^D$ to be represented by an anchor point $\{\mathbf{c}_k\}, k\in K$ which is achieved by dividing the descriptor space $\mathbb{R}^D$ into $K$ cells using a vocabulary of $K$ person words. The VLAD aggregation records counts of visual words by storing the sum of residuals (difference vector between the local descriptor $\mathbf{f}_b$ and its corresponding cluster centre $\mathbf{c}_k$) for each visual word $\mathbf{c}_k$.

In our case, given $B$ local convolutional feature cubic $\{\mathbf{f}_b\}$ as inputs, and $K$ cluster centres (a.k.a visual words) $\{\mathbf{c}_k\}$ as the VLAD parameters, the output of VLAD aggregation is the representation matrix $\mathbf{V} \in K\times D$. The element of $\mathbf{V} [j,k]$ is computed as
\begin{equation}\label{eq:V-matrix}
  \mathbf{V} [j,k]=\sum_{b=1}^B a_k( \mathbf{f}_b ) ( f_b(j)-c_k(j)),
\end{equation}
where $f_b(j)$ and $c_k(j)$ are the $j$-th dimensions of the $b$-th feature and $k$-th cluster, respectively. $a_k( \mathbf{f}_b )$ denotes the membership assignment of $\mathbf{f}_b$ to $k$-th visual word, \ie $a_k( \mathbf{f}_b)=1$ if $\mathbf{c}_k$ is the closest cluster to $\mathbf{f}_b$, and $a_k(\mathbf{f}_b)=0$ otherwise. Thus, each column $k$ of $\mathbf{V}$ records the sum of residuals $(\mathbf{f}_b -\mathbf{c}_k)$ of features which are assigned to cluster $\mathbf{c}_k$.

However, the computation of VLAD is not continuous. Designing a trainable generalized VLAD layer plug into CNNs requires the operation is differentiable w.r.t all its parameters and the input. To make the VLAD aggregation differentiable, the hard assignment $a_k(\mathbf{f}_b)$ of $\mathbf{f}_b$ to $\mathbf{c}_k$ is replaced with soft assignment to multiple clusters \cite{NetVLAD}, that is, $\hat{a}_k(\mathbf{f}_b)=\frac{ e^{-\alpha ||\mathbf{f}_b-\mathbf{c}_k||^2} }{ \sum_{k'} e^{-\alpha ||\mathbf{f}_b-\mathbf{c}_{k'}||^2}}$, which assigns the weight of $\mathbf{f}_b$ to cluster $\mathbf{c}_k$ proportional to their proximity. $\alpha$ is a tunable hyper-parameter, and when $\alpha \rightarrow \infty$, $\hat{a}_k(\mathbf{f}_b)$ turns to be the original VLAD. Hence, each video descriptor $\mathbf{f}_b$ is assigned to one of the cells and represented by a residual vector $\mathbf{f}_b-\mathbf{c}_k$. The difference vectors are then summed across the entire temporal extent as
\begin{equation}\label{eq:residual-sum}
  \mathbf{V}[j,k]=\sum_{b=1}^B \underbrace{ \frac{ e^{-\alpha ||\mathbf{f}_b-\mathbf{c}_k||^2} }{ \sum_{k'} e^{-\alpha ||\mathbf{f}_b-\mathbf{c}_{k'}||^2}} }_{Soft-asignment} (\underbrace{f_b(j) -c_k (j)}_{Residual} ).
\end{equation}

In Eq.\eqref{eq:residual-sum}, the first term represents the soft-assignment of descriptor $\mathbf{f}_b$ to cell $k$ and the second term of $f_b(j) -c_k(j)$ is the residual between the descriptor and the anchor point of cell $k$. The summing operation over all regions is able to aggregate all local features over the entire video because the 3D ConvNets are designed to be convolutional sampling across all frames with a spatiotemporal kernel. The output of the aggregation layer is a matrix $\mathbf{V}$, where the $k$-th column $\mathbf{V}[\cdot,k]$ represents the aggregated descriptors in the $k$-th cell. The columns of the matrix are then intra-normalized \cite{All-VLAD}, stacked and L2-normalized \cite{VLAD} into a single descriptor $\mathbf{v}\in \mathbb{R}^{KD}$ of the entire video sequence.

Intuitively, by expanding the Eq. \eqref{eq:residual-sum} and removing the term $e^{-\alpha}||\mathbf{f}_b||^2$, we have the soft-assignment in the following form:
\begin{equation}\label{eq:new-soft}\small
  \hat{a}_k(\mathbf{f}_b)=\frac{ e^{\mathbf{w}_k^T \mathbf{f}_b + z_k} }{ \sum_{k'} e^{\mathbf{w}_{k'}^T \mathbf{f}_b + z_{k'} } }, \mathbf{V}[j,k]=\sum_{b=1}^B \frac{ e^{\mathbf{w}_k^T \mathbf{f}_b + z_k} }{ \sum_{k'} e^{\mathbf{w}_{k'}^T \mathbf{f}_b + z_{k'} } } (f_b(j) -c_k (j)),
\end{equation}
where the independent vectors $\{\mathbf{w}_k\}= 2\alpha \{\mathbf{c}_k\}$, scalar $z_k=-\alpha ||\mathbf{c}_k||^2$, and $\{\mathbf{c}_k\}$ are sets of trainable parameters for each cluster $k$. The anchor point $\mathbf{c}_k$ can be interpreted as the origin of a new coordinate system local to the cluster $k$. Hence, all components in the proposed model including convolutional feature extractor, attention-based part alignment, person words, and the classifier are differentiable operations. Also, the spatiotemporal aggregation and the following L2-normalization are differentiable and allow for back-propagating error gradients to lower layers of the network. The soft deterministic attention mechanism on different body parts is also differentiable \cite{ShowAttendTell}. As all components are differentiable, one can back-propagate through the network architecture to joint learn the optimal weights for all parameters from the video data in an end-to-end manner so as to better discriminate the target person. Thus, the proposed model \textbf{3D PersonVLAD} is designed to aggregate 3D body part features into a compact video-level representation across the entire video extent.

\begin{figure}[t]
\centering
\includegraphics[width=9cm,height=2.5cm]{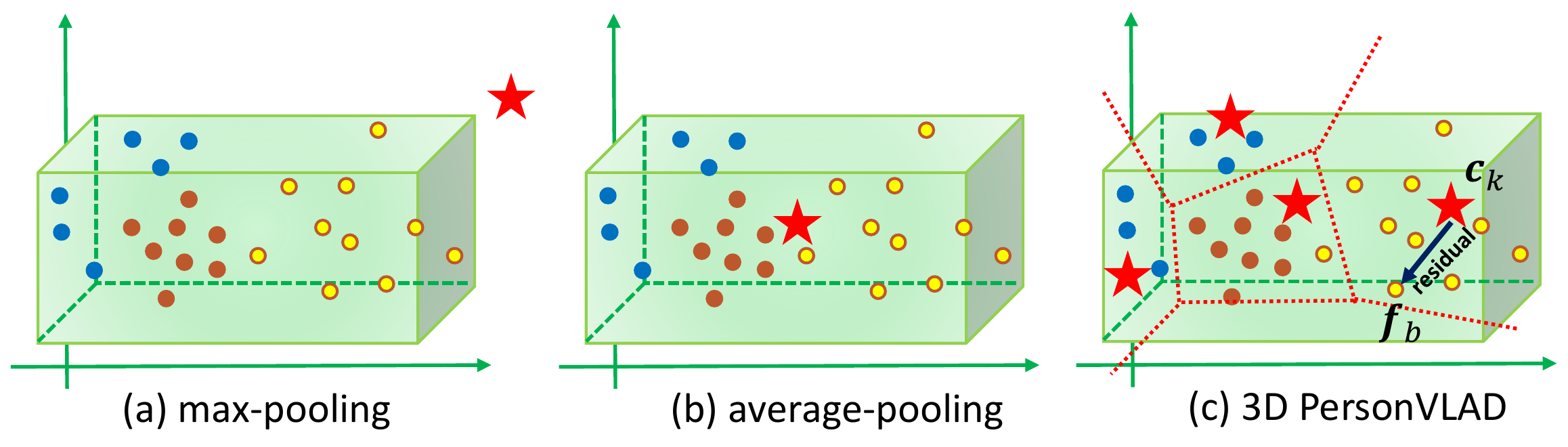}
\caption{Different pooling strategies for a collection of divergent feature descriptors. Points represent features from a video and different colors respond to different motion cycles of persons. The max or global/average pooling are suitable for similar features, however they do not capture the whole feature distribution space. The PersonVLAD representation effectively clusters the appearance and motion features and aggregates their residuals from the nearest cluster centers.}\label{fig:different-pooling}
\end{figure}

\paragraph{Discussions on Other Pooling Operations}
It is worth remarking the differences of the above aggregation compared to alternative pooling strategies: 3D global/average pooling and the common max pooling. The 3D global pooling can also be applied to produce video-level features, as can be performed over the potential regions (see Fig.\ref{fig:3d-parts}). With a global/average pooling over all locations across frames, we can have $\bar{\mathbf{f}}_b=AvePooling(\mathbf{F}_b)$, where $\bar{f}_b(c)=Average_{l,x,y}[\mathbf{f}_b(c,l,x,y)]$. The global average pooling is to not only minimize the over-fitting by reducing the number of parameters in the model but also augment the localization ability of the network. Thus, each of the activation maps in the $B$ branches can act as detector for an attentive region against various variations so as to address the misalignment of body parts. Thereafter, features from detected regions are extracted (i.e., $\bar{\textbf{f}}_b$), followed by a fully-connected layer to transform $\bar{\mathbf{f}}_b$ to a $d$-dim feature vector $\mathbf{f}_b=\mathbf{W}^7_{FC_b}(\mathbf{W}^6_{FC_b}\bar{\mathbf{f}}_b)$, with a non-linear ReLU operation subsequently. These fully-connected layers on each branch do not share parameters because different body regions should have different importance when used for identifying persons. Thus, the embedding on $B$ parts should be optimized separately. Finally, all part features are concatenated to produce the clip-level feature output. Thus, we have $\mathbf{f}=[\mathbf{f}_1^T, \mathbf{f}_2^T,\ldots, \mathbf{f}_B^T]^T$, which is used to be the pedestrian representations. Similarly, the max-pooling is in the form $\bar{\mathbf{f}}_b=MaxPooling(\mathbf{F}_b)$, the transformation to form the video representation is the same as average pooling. However, the global/average or max pooling represents the entire distribution of points as only a single descriptor which can be sub-optimal for representing an entire video composed of multiple person motion cycles. In contrast, the proposed video aggregation represents an entire distribution of descriptors with multiple motion cycles by splitting the descriptor space into cells and pooling inside each of the cells. Moreover, the proposed residual aggregation which can be interpreted as a strong regularization constraint, is more suitable for training on limited person re-ID data while producing a global descriptor with a high dimensionality of $KD=16,384$. The comparison on different pooling operations is shown in Fig.\ref{fig:different-pooling}.

\subsection{Training Procedure}\label{ssec:training}

To learn the global discriminative representation, we train our 3D networks on MARS \cite{MARS} which is currently the largest video benchmark for person re-ID. The training is done on the MARS train split containing 5K bounding boxes for 625 identities. We adopted a two-step training procedure. First, given the competitive performance of 3D ConvNets on video classification tasks, we use C3D \cite{3DConvNets} as the initialization for all the 3D convolutions and pooling which is training the network on Sports-1M train split \cite{3DConvNets}. Then, the parameters of 3D part alignment module and the classifier are random initialized. In the aggregation layer, we use $K=64$ and a high value for $\alpha=1000.0$, as suggested in \cite{ActionVLAD,NetVLAD}. The $\alpha$ parameter is initially chosen to be large such that the soft assignment weights are sparse to resemble the conventional VLAD. To simplify the learning, we decouple the parameters $\{\mathbf{c}_k\}$ in the aggregation layer that are used to compute the soft assignment and the residual in Eq.\eqref{eq:residual-sum}. Decoupling parameters is a means to adapt VLAD into a new dataset \cite{All-VLAD}. Pedestrian labels are regarded as training target and identification loss (Eq.\eqref{eq:loss}) is used to fine-tune these parameters.

In the training, all video frames are fixed into the size of $90\times 180$ and split into 16-frame clips to fit into the networks. Thus, the input dimensions are $3\times 16\times 90\times 180$. The input frames are downsized in order to keep a larger batch size (e.g., 15 in our training) to reduce the batch bias so as to achieve faster  and stable convergence. Unlike \cite{3DConvNets}, we do not perform random cropping on the input clips because this spatial/temporal jittering would cause visual loss in the visual appearance of persons. We also horizontally flip them with 50\% probability. The training batch size is 15. All convolution layers are applied with appropriate padding (both spatial and temporal) and the stride is 1. The kernel temporal depth is $d=3$ throughout all convolutions in accordance to the best setting of C3D. When training the PersonVLAD, we use the Adam solver \cite{ADAM} with $\epsilon=10^{-4}$. This is required as the PersonVLAD output is L2-normalized and a lower value of $\epsilon$ is needed to ensure a fast convergence. Thus, we perform a two-step training. First, we train the network from scratch but initialize and fix the PersonVLAD cluster centers, and only train the classifier with a learning rate of 0.003. The learning rate is divided by 2 after very 1K iterations. The optimization is stopped at 50K iterations. In the second step, we jointly fine tune the other components and the PersonVLAD cluster centers with a smaller learning rate of $10^{-4}$.

To predict the person's identity, we follow the previous works \cite{Diversity-Video-Re-ID,OIM} and use the Online Instance Matching loss function (OIM) as the objective loss. The identification loss is defined as follows:
\begin{equation}\label{eq:loss}
p_i =\frac{\exp(e_i^T \mathbf{v} \diagup \tau )}{ \sum_{j=1}^{\mathbb{C}}\exp(e_j^T \mathbf{v}\diagup\tau) + \sum_{k=1}^{\mathbb{Q}}\exp(\mathbf{u}_k^T \mathbf{v}\diagup\tau)},
\end{equation}
where $e_i$ is the $i$-th column (corresponding to the $i$-th labeled identity) of the lookup table $\mathbf{E}$ that stores the features of all the labeled identities. $\mathbb{C}$ denotes the total number of labeled identities. $\mathbf{U}$ denotes the features in the circular queue with queue size $Q$ that stores the features of these unlabeled identities that appear in the recent mini-batches. The temperature parameter $\tau$ controls the softness of probability distribution (e.g. a higher value $\tau$ leads to a softer probability distribution. In our experiments, the temperature scalar is set to be 0.1 and $Q$ is set to be 5,000 according to the practice \cite{Diversity-Video-Re-ID} \footnote{The probability of being classified as the $i$-th unlabelled identity in $Q$ is defined as $q_i =\frac{\exp(u_i^T \mathbf{v} \diagup \tau )}{ \sum_{j=1}^{\mathbb{C}}\exp(e_j^T \mathbf{v}\diagup\tau) + \sum_{k=1}^{\mathbb{Q}}\exp(\mathbf{u}_k^T \mathbf{v}\diagup\tau)}$}). Intuitively, this OIM loss is optimized be able to minimize the features discrepancy among the instances of the same person, while maximize the discrepancy among different people by memorizing the features of all the people. Specifically, the OIM loss function uses a lookup table to store features of all identities appearing in the training set. In each forward iteration, a mini-batch sample is compared against all the identities when computing classification probabilities. This loss function has shown to be very more effective than softmax loss in training re-ID networks \cite{OIM}.

\subsection{Visualization of Features}\label{ssec:visualization}

\begin{figure*}[t]
\centering
\includegraphics[width=13cm,height=3cm]{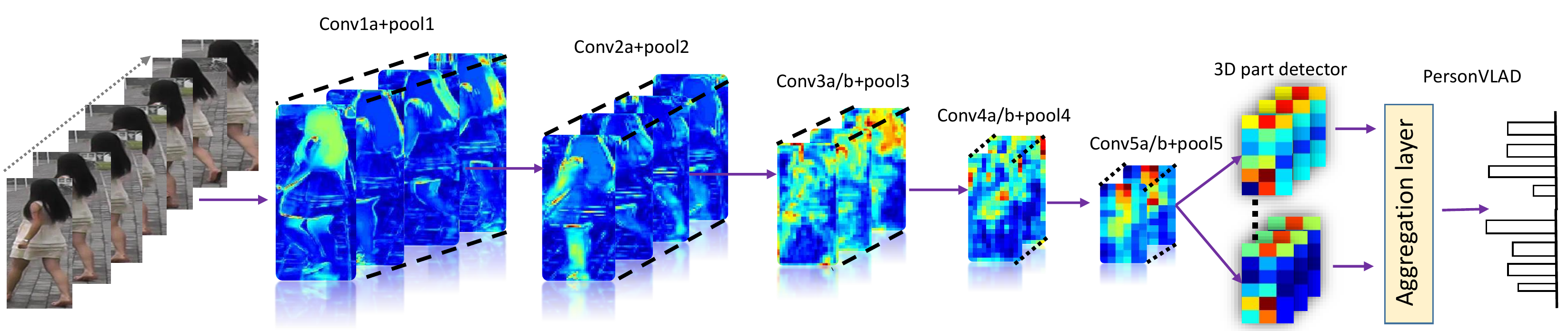}
\caption{Visualization of features learned by our network. Initial layers tend to learn spatiotemporal features evolving over time while deeper layers learn to localize distinct regions so that the extracted local features are highly discriminative to be aggregated into global representation and the classification performance is maximized. Refer to Section \ref{ssec:visualization} for more details.}\label{fig:visualize-features}
\end{figure*}

Fig.\ref{fig:visualize-features} shows feature response visualization at each layer of the 3D person re-ID network. The response maps of Conv1a/pool1 show the responses after Conv1a but before poo1. The features respond strongly to the black regions of the input sequence, highlighting the hair regions. This is mainly because the 3D filter can encapsulate the spatial dimension and temporal dependency during convolutions. The features after Conv2a indicate that after the second 3D convolution, the filters capture tan and skin-color regions, giving high responses to the legs and hands of the person. Since it is 3D filter, similar parts of the sequence are highlighted across frames. After another three 3D convolution and pooling, a 3D part module with $B$ branches is applied to detect $B$ distinct regions that are discriminative to the person. The resultant local features from different parts are passed through the PersonVLAD layer, followed by a L2-normalization to produce the global video-level representation.

%% file: experiment.tex
\section{Experiments}\label{sec:exp}

In this section, we present empirical evaluations on the proposed model. Experiments are conducted on three benchmark video sequences for person re-identification.

\begin{figure*}[t]
\centering
\begin{tabular}{c}
\includegraphics[height=2cm,width=12cm]{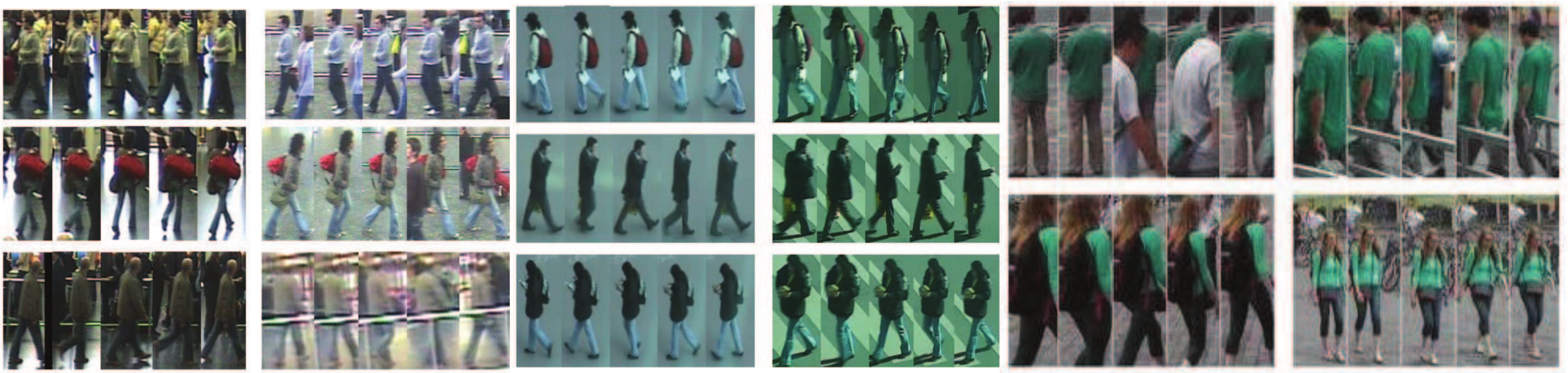}
\end{tabular}
\caption{Left: iLIDS-VID. Middle: PRID2011. Right: MARS. Image sequences of the same pedestrian (in row) in different camera views from the three datasets.}
\label{fig:examples}
\end{figure*}

\subsection{Datasets and Evaluation Metrics}

We validate our method and compare to state-of-the-art approaches on three data sets (Fig.\ref{fig:examples}): MARS \cite{MARS}, iLIDS-VID \cite{VideoRanking}, and PRID 2011 \cite{PRID2011}.

\begin{itemize}
\item The MARS dataset contains 1,261 pedestrians, and 20,000 video sequences, making it the largest video re-ID dataset. Each sequence is automatically obtained by the Deformable Part Model \cite{DetectionPAMI} detector and the GMMCP \cite{GMMCP} tracker. These sequences are captured by six cameras at most and two cameras at least, from which each identity has 13.2 sequences on average. This dataset is evenly divided into train and test sets, containing 625 and 636 identities, respectively.
\item The iLIDS-VID dataset consists of 600 image sequences for 300 randomly sampled people, which was created based on two non-overlapping camera views from the i-LIDS multiple camera tracking scenario. The sequences are of varying length, ranging from 23 to 192 images, with an average of 73. This dataset is very challenging due to variations in lighting and viewpoint caused by cross-camera views, similar appearances among people, and cluttered backgrounds.
\item The PRID 2011 dataset includes 400 image sequences for 200 persons from two adjacent camera views. Each sequence is between 5 and 675 frames, with an average of 100. This dataset was captured in uncrowded outdoor scenes with rare occlusions and clean background. Also, the dataset has obvious color changes and shadows in one of the views.
\end{itemize}

A person re-ID system can be tested as a ranking problem where a probe/query in camera 1 is issued to compute its similarity against each candidate in the gallery set under a different camera 2. The expectation is that the correct matched candidates to the probe pedestrian in camera 2 will be ranked at the top. The widely used evaluation metric is Cumulative Matching Characteristics (CMC) \cite{CMC}, which is an estimate of the expectation of finding the correct match in the first $n$ matches. We also report the Mean Average Precision (mAP) \cite{Market1501} over MARS.

\begin{table}[t]\small
  \centering
  \caption{Statistical characteristics of three datasets.} \label{tab:datasets}
  {
  \begin{tabular}{l|c|c|c}
  \hline
    Datasets  & iLIDS-VID  &  PRID2011  & MARS\\
  \hline
  $\sharp$identities & 300 & 200 &1,261\\
  $\sharp$tracklets & 600 & 400 & 20,478 \\
  $\sharp$bboxes & 44K & 40K & 1M\\
  $\sharp$distractors & 0 & 0 & 3,248\\
  $\sharp$cameras & 2 & 2 & 6\\
  $\sharp$resolution & $64\times 128$ & $64\times 128$ & $128\times 256$\\
  $\sharp$detection & hand & hand & DPM+GMMCP\\
  $\sharp$evaluation & CMC & CMC & mAP+CMC\\
  \hline
  \end{tabular}
  }
\end{table}

\subsection{Empirical Analysis}\label{ssec:analysis}
In this section, we conduct extensive empirical evaluations to analyze the properties of the proposed network in terms of the following aspects. All empirical experiments are conducted on the MARS dataset.

\paragraph{Number of Tracklets for Video Descriptors}
Once training is finished, the proposed 3D person net can be used to extract video-level descriptor for each identity in test. In the test split of MARS, the number of tracklets regarding most individual varies from 5 to 20 (average 13.2), and thus, it is necessary to determine how many tracklets should be selected for feature extraction. To this end, we study the effect of the selection of tracklets with appropriate pooling.

\begin{itemize}
\item Random tracklet (probe) and all tracklets (gallery):  For each probe identity, a tracklet is randomly selected whilst all tracklets of the gallery candidates are used. To extract 3D features, each tracklet is split into 16 frame long clip with a 8-frame overlap between two consecutive clips. These clips are put through the network to produce its global feature. These clip-level activations are averaged to form a final 16,384-dim video descriptor, namely \textit{tracklet activation}. This process is performed 10 times and the CMC results are averaged.
\item All tracklets (probe) and all tracklets (gallery): For each probe/gallery identity, all tracklets are used to produce their tracklet activations which are averaged to form a global vector.
\item Random tracklet (probe) and random tracklet (gallery): For each probe/gallery identity, a tracklet is randomly selected to obtain the tracklet activations. It is performed 10 times and the CMC results are  averaged.
\item All tracklets (probe) and random tracklet (gallery): In this setting, for each gallery identity, a tracklet is randomly selected whilst all tracklets of the probes are used to extract their video descriptors.
\end{itemize}

Table \ref{tab:tracklets} shows the CMC values against varied selection of tracklets in probe and gallery. It can be seen that multiple tracklets in the gallery are beneficial to the re-ID recognition whilst increased tracklets for the probe are not helpful. For example, a random tracklet of probe against all tracklets of gallery can yield rank-1=80.8 while all tracklets for both probe and gallery achieves the similar result (rank-1=80.2). To balance the complexity, we refer to this representation (random tracklet of probe and all tracklets of gallery) as 3D global video descriptor in all experiments unless we specify the difference.

\begin{table}[t]\small
  \centering
  \caption{The effect of varied tracklets for probe and gallery on MARS. The CMC values of rank-1 are reported.} \label{tab:tracklets}
  {
  \begin{tabular}{l|c|c}
  \hline
    \diaghead{GalleryProbe}{Gallery}{Probe}  & random  (CMC@1) & all (CMC@1) \\
  \hline
  random (CMC@1) & 74.1 & 71.7\\
  all (CMC@1) & 80.8 & 80.2 \\
  \hline
  \end{tabular}
  }
\end{table}

\paragraph{Number of Body Parts and Spatial Partition}

\begin{figure}[t]
\centering
\includegraphics[width=5cm,height=4cm]{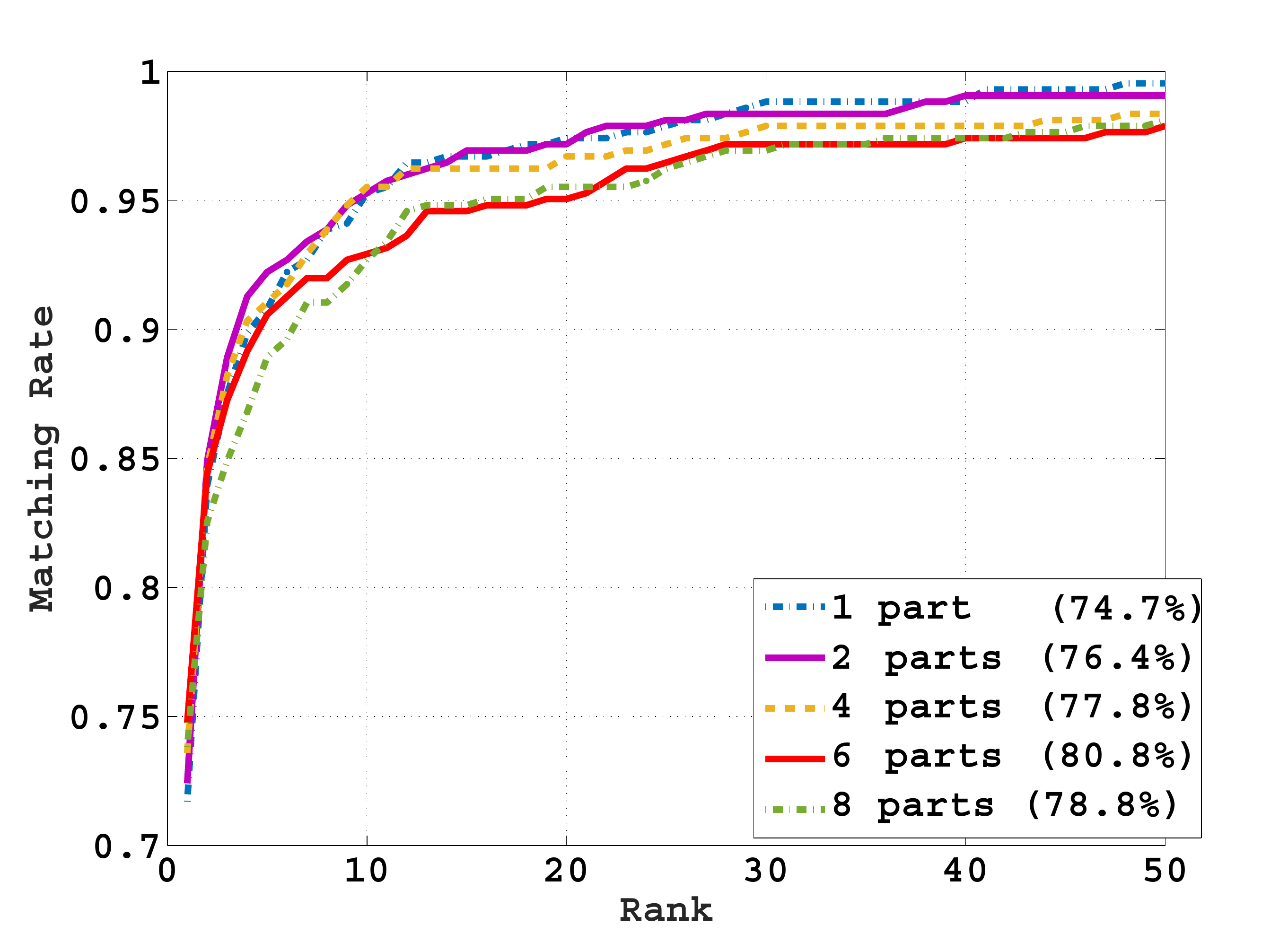}
\caption{The effect of varied number of body parts on MARS.}\label{fig:body-parts}
\end{figure}

\begin{table}[t]\small
  \centering
  \caption{The comparison of our approach and spatial partition based methods (stripe and fixed grid) over MARS.} \label{tab:spatial-partition}
  {
  \begin{tabular}{c|c|c|c|c|c}
  \hline
    Datasets  & Method  &  rank-1 & rank-5 & rank-10 & rank-20\\
  \hline
\multirow{4}{*}{MARS} & stripe & 68.3 & 87.2 &90.0 & 94.6\\
& grid & 67.1 & 86.0 & 89.5 & 94.0\\
& 3D ConvNets+GP & \color{red}$\mathbf{74.8}$ & \color{red}$\mathbf{90.6}$ & \color{red}$\mathbf{92.9}$ & \color{red}$\mathbf{95.1}$ \\
& PersonVLAD & \color{red}$\mathbf{80.8}$ & \color{red}$\mathbf{94.5}$ & \color{red}$\mathbf{96.9}$ & \color{red}$\mathbf{99.0}$ \\
  \hline
  \end{tabular}
  }
\end{table}

\begin{table}[t]\small
  \centering
  \caption{The validation with varied numbers of body parts ($B$) over MARS. The network is trained on a random half of the training data and the validation is performed on the remaining half.} \label{tab:number-parts}
  {
  \begin{tabular}{c|c|c|c|c}
  \hline
    No. parts  &  rank-1 & rank-5 & rank-10 & rank-20\\
  \hline
1 & 74.7 & 91.8 & 94.3 & 96.4\\
2 & 76.4 & 92.2 & 94.3 & 97.2\\
4 & 77.8 & 92.4 & 95.5 & 98.7\\
6 & \color{red}$\mathbf{80.8}$ & \color{red}$\mathbf{94.5}$ & \color{red}$\mathbf{96.9}$ & \color{red}$\mathbf{99.0}$ \\
8 & 78.8 & 91.9 & 95.7 & 98.2 \\
  \hline
  \end{tabular}
  }
\end{table}

\begin{figure}[t]
\centering
\includegraphics[width=8.5cm,height=4cm]{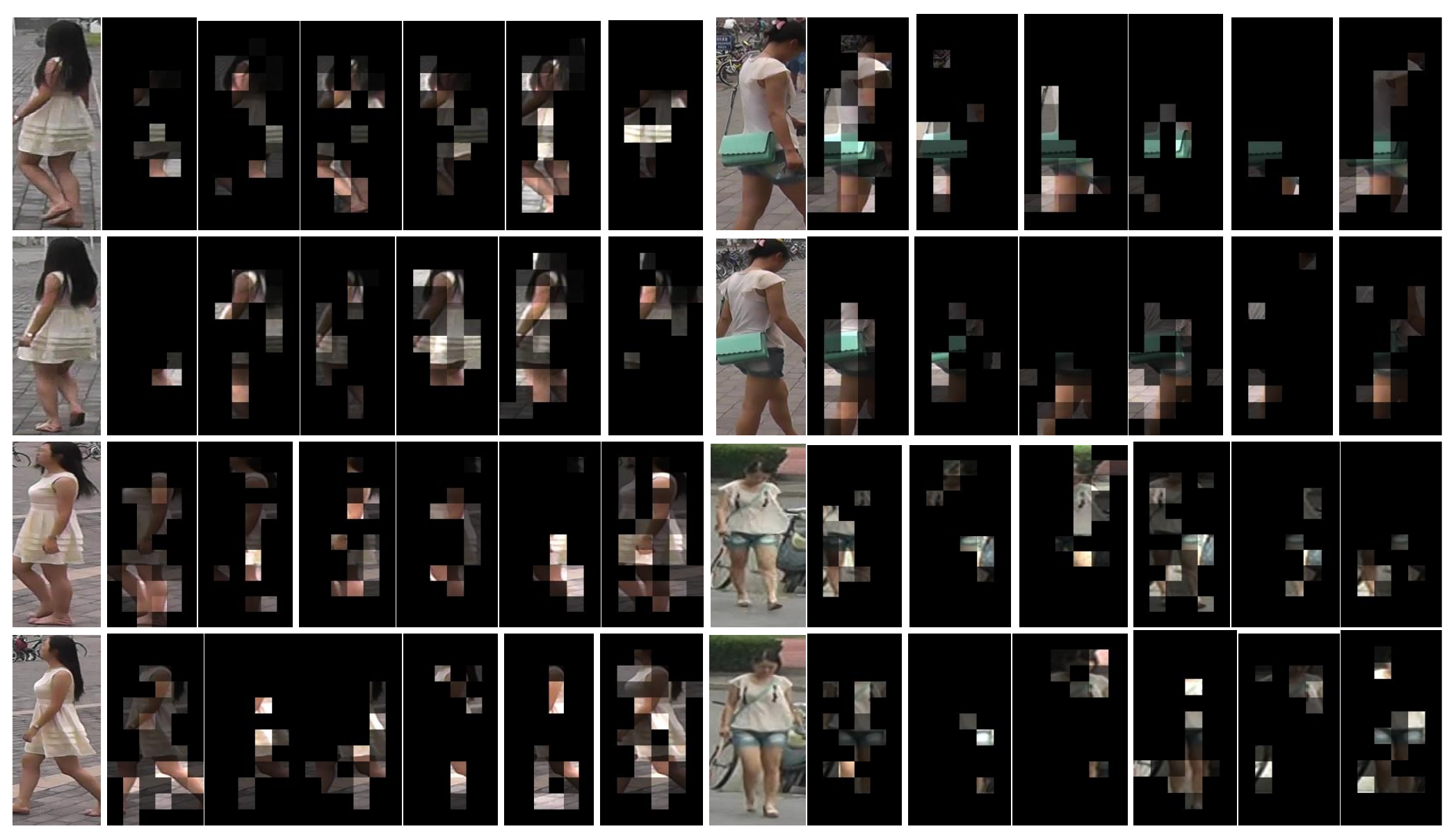}
\caption{Illustration of the part maps learned by the 3D part alignment module. For each frame box, our method can generate 6 part maps corresponding to distinct regions.}\label{fig:attention-parts}
\end{figure}

To determine the number of body parts ($B$), we quantitatively evaluate the our method with respect to varied number of body parts. Table \ref{tab:number-parts} and Fig.\ref{fig:body-parts} show the results associated with varied number of body parts from 1, 2, 4, 6, and 8. The empirical evaluation indicates that 6 body partition is more suitable for MARS in which pedestrians exhibit diverse visual appearance and finer body partition is needed. Fig.\ref{fig:attention-parts} shows the attentive parts our approach learns through the 3D part-based alignment. It can be seen that body parts of pedestrians under disjoint camera views are generally well aligned for the same identity. For example, the parts almost describe the same regions on different frames.

To study how well the misalignment is addressed in the learned part based representations, we empirically compare our approach with two alternative spatial partition schemes: dividing the frame box into 5 horizontal stripes or $5\times 5$ grids to form the region maps. The new region maps are used to replace the part masks in our approach and train the network to produce spatial partition-based features. The results given in Table \ref{tab:spatial-partition} demonstrate that local body part partition is more effective than spatial partition alternatives. This is mainly because the pose changes or uncontrolled human spatial distributions in the human frame box, and pre-defined spatial partition such as stripe or grid-based, may not be well-aligned with human body parts. In contrast, the proposed 3D local features encapsulate the distinct regions and address the misalignment in their representation learning stage. Based on the 3D part alignment module, the aggregation layer can well preserve the identity discrimination across the entire video span. Thus, it enables high similarity search values with the Euclidean distance.

\paragraph{Comparison with Motion Features}

In this experiment, we compare the learned deep 3D global features with different motion features below.

\begin{itemize}
\item \textbf{HOG3D} extracts 3D HOG features \cite{HOG3D} from volumes of video data \cite{VideoRanking}. After extracting a walk cycle by computing local maxima/minima of the Flow Energy Profile (FEP) signal, video fragments are further divided into $2\times 5$ (spatial) $\times$ 2 (temporal) cells with 50\% overlap. A spatiotemporal gradient histogram is computed for each cell which is concatenated to form the HOG3D descriptor.

\item \textbf{STFV3D} is a low-level feature-based Fisher vector learning and extraction method which is applied to spatially and temporally aligned video fragments \cite{VideoPerson}. STFV3D proceeds as follows: 1) temporal segments are obtained separately by extracting walk cycles \cite{VideoRanking}, and spatial alignment is implemented by detecting spatial bounding boxes corresponding to six human body parts; 2) Fisher vectors are constructed from low-level feature descriptors on those body-action parts.

\item \textbf{C3D} features \cite{3DConvNets} are state-of-the-art deep spatiotemporal features for videos. We use the pre-trained model and fine-tune the parameters on MARS. The input video clips have 16 frames and the network is set up to take video clips as inputs and predict the class labels belonging to 625 identities.
\end{itemize}

Table \ref{tab:motion-features} reports the comparison results yielded by different motion features. We can see that low-level features of HOG3D and STFV3D have low matching rate on MARS. The primary reason is in a large dataset there are many pedestrian sharing similar motion features w.r.t the probe. Thus, it is difficult to discriminate different persons based on their motions. Also, since MARS has six cameras to capture videos of persons in which the same identity may exhibit dramatic variations due to pose and viewpoint changes. Thus, spatial misalignment is a significant issue to be addressed and STFV3D yields under-performed matching rates otherwise. Deep C3D features are able to capture the visual appearance changes and motions more precisely and yield promising recognition results. However, C3D features still learn global features to describe persons whilst human body regions across cameras are not well aligned. In contrast, our 3D global features are learned to not only capture spatial and temporal dynamics simultaneously but also localize body regions and extract local features to form global discriminative representations.

\begin{table}[t]\small
  \centering
  \caption{Comparison results of different motion features.} \label{tab:motion-features}
  {
  \begin{tabular}{l|c|c|c}
  \hline
    Features  &  rank-1 & rank-5  & rank-20\\
  \hline
HOG3D \cite{HOG3D}+KISSME \cite{KISSME} & 2.6 &6.4  &12.4\\
STFV3D \cite{VideoPerson}+ KISSME \cite{KISSME} & 22.0& 33.4 & 59.0\\
C3D \cite{3DConvNets}  & 58.9 & 81.0  & 93.7 \\
PersonVLAD & \color{red}$\mathbf{80.8}$ & \color{red}$\mathbf{94.5}$ & \color{red}$\mathbf{99.0}$ \\
  \hline
  \end{tabular}
  }
\end{table}

\paragraph{Cross-dataset Evaluation}

In this experiment, we study the generalization capability of our model across data sets. This is conducted by transferring the optimized parameters from MARS to two smaller target data sets: iLIDS-VID and PRID2011. First, we directly transfer the model trained on MARS to iLIDS-VID and PRID2011, denoted as MARS$\rightarrow$ iLIDS-VID (PRID2011). In Table \ref{tab:cross-dataset}, it shows that directly extracting features pre-trained on MARS for iLIDS-VID and PRID2011 cannot obtain high re-id accuracy due to the disparity of data samples between MARS and the target data sets. The second trail is fine-tuning the MARS-trained model on the training splits of the two targets. For both data sets, a random half of the data sets are divided into training and the rest half are used as testing. On both data sets, fine-tuning yields notable improvement over the direct model transferral. On iLIDS-VID, we achieve rank-1=69.4, which is much higher than that value of MARS$\rightarrow$ iLIDS-VID (rank-1=56.3). On PRID2011, fine-tuning strategy is higher than MARS $\rightarrow$ PRID2011 by 16\% at rank-1.

\begin{table}[t]\small
  \centering
  \caption{Cross-dataset evaluation results.} \label{tab:cross-dataset}
  {
  \begin{tabular}{l|c|c|c}
  \hline
    Training  &  rank-1 & rank-5  & rank-20\\
  \hline
 MARS $\rightarrow$ iLIDS-VID & 56.3  & 78.2 &93.7\\
 Fine-tune on iLIDS-VID & 69.4 & 87.6& 99.2\\
 MARS$\rightarrow$ PRID2011 & 71.7 & 85.6 & 94.1\\
 Fine-tune on PRID2011  & 87.6 & 96.1 & 98.7 \\
  \hline
  \end{tabular}
  }
\end{table}

%
%

\subsection{Comparison with CNN Video Representations}\label{ssec:compare-ConvNets}
Recent works in learning video representations directly from data using CNNs can be categorized into two-stream architectures and spatiotemporal convolutions. In this experiment, we compare the proposed PersonVLAD aggregation with a variety of CNN video features learned by different architectures with varied aggregation options.

\paragraph{Two-steam ConvNets}

\begin{table}[t]
\caption{Comparison with state-of-the-art two-stream ConvNets.}\label{tab:com-two-stream}
\centering
\begin{tabular}{l|c|c|c}
\hline
Dataset & iLIDS-VID & PRID2011 & MARS\\
\hline
Rank @ R & R = 1  & R = 1   & R = 1  \\
\hline
VGG-16-Element-Max & 56.2& 71.6 & 65.7\\
VGG-16-Element-Average & 57.1& 73.8& 65.9\\
VGG-16-Element-Multiplication & 60.4& 76.3& 67.8\\
Two-stream Fusion \cite{Two-stream-fusion} & 64.1 & 83.0 &74.9\\
ActionVLAD \cite{ActionVLAD} & 65.6 & 83.7 &76.5\\
FstCN \cite{FstCN} & 59.1 & 76.8 &68.3\\
VideoDarwin \cite{VideoDarwin} & 62.5 & 81.1 & 73.4\\
RNN+FV \cite{RNN-FV} & 58.4 & 78.3 &70.2\\
PersonVLAD & 69.4 & 87.6 & 80.8 \\
\hline
\end{tabular}
\end{table}

In our evaluation, we consider three two-stream architecture baselines which are varied on different aggregation functions: element-wise maximum, element-wise average and element-wise multiplication. To this end, we employ the VGG-16 \cite{VGG} for the design of two-stream ConvNets, which consist of spatial and temporal networks. The spatial ConvNet operates on the RGB frames, and the temporal ConvNet operates on a stack of 10 dense optical flow frames. The input RGB frames are resized into $224\times 224$, and then mean-subtracted for network training. To fine-tune the network, the original classification layer is replaced with a $\mathtt{C}$-way softmax layer where $\mathtt{C}$ indicates the number of training identities. The prediction scores of the spatial and temporal ConvNets are combined in a late fusion manner as averaging before softmax normalization.

Other state-of-the-art are also considered including Two-stream Fusion \cite{Two-stream-fusion}, ActionVLAD \cite{ActionVLAD}, FstCN \cite{FstCN}, VideoDarwin \cite{VideoDarwin}, and RNN+FV \cite{RNN-FV}. The comparison results are shown in Table \ref{tab:com-two-stream}. We observe that amidst the two-stream baselines, the element-wise multiplication performs the best. This observation suggests a feasible way of aggregating the appearance and motion information, however, element-wise multiplication leads to a high-dimensional feature, and thus not efficient to compute. In comparison with the current methods using two-stream ConvNets, PersonVLAD performs the best among all methods. The performance gap between PersonVLAD and ActionVLAD \cite{ActionVLAD} is 3.8\%, 3.9\%, and 4.3\% on the three benchmarks. Also, PersonVLAD is 5.3/4.6/5.9\%, 10.3/10.8/12.5, 6.9/6.5/7.4\%, 11/9.3/10.6\% better than Two-stream Fusion \cite{Two-stream-fusion}, FstCN \cite{FstCN}, VideoDarwin \cite{VideoDarwin}, and RNN+FV \cite{RNN-FV} methods on the three benchmarks. Hence, our model clearly show the effectiveness of encoded feature representation in video-level for entire videos, in end-to-end learning. Moreover, our PersonVLAD has fewer parameters to train, in comparison to other methods which have several fully-connected layers to train, such as Two-stream Fusion \cite{Two-stream-fusion}.

\paragraph{3D ConvNets}

\begin{table}[t]
\caption{Comparison with state-of-the-art 3D ConvNets.}\label{tab:com-3D}
\centering
\begin{tabular}{l|c|c|c}
\hline
Dataset & iLIDS-VID & PRID2011 & MARS\\
\hline
Rank @ R & R = 1  & R = 1   & R = 1  \\
\hline
C3D \cite{3DConvNets} & 60.4 & 73.8 & 58.9\\
LTC \cite{LTC} & 61.7 & 75.2 & 67.4 \\
iDT+FV \cite{iDT} & 60.8 & 76.7 & 59.8\\
LRCN \cite{LRCN} & 59.5 & 73.0 & 58.1\\
Spatio-Temporal ConvNet \cite{Karpathy} & 58.1 & 68.4 & 51.6\\
Composite LSTM \cite{VideoLSTM} & 54.8 & 65.4 & 53.1\\
SpaAtn+TemAtn \cite{Diversity-Video-Re-ID} & 69.7 & 88.4 & 77.1 \\
PersonVLAD & 69.4 & 87.6 & 80.8 \\
\hline
\end{tabular}
\end{table}

In Table \ref{tab:com-3D}, we compare the performance of 3D PersonVLAD with state-of-the-art 3D ConvNets which include C3D \cite{3DConvNets}, LTC \cite{LTC}, iDT+FV \cite{iDT}, LRCN \cite{LRCN}, Spatio-Temporal ConvNet \cite{Karpathy}, and Composite LSTM \cite{VideoLSTM}. Similar to two-stream ConvNets, the proposed 3D PersonVLAD outperforms other methods, and achieve an accuracy of 69.4\%, 87.6\%, and 80.8\% on iLIDS-VID, PRID2011, and MARS, respectively, which is 9/8.6\%, 13.8/10.9\%, and 21.9/21\% better than the original C3D ConvNet \cite{3DConvNets}, and iDT+FV \cite{iDT} methods on the three data sets.

\begin{figure*}[t]
\centering
\begin{tabular}{ccc}
\includegraphics[height=4cm]{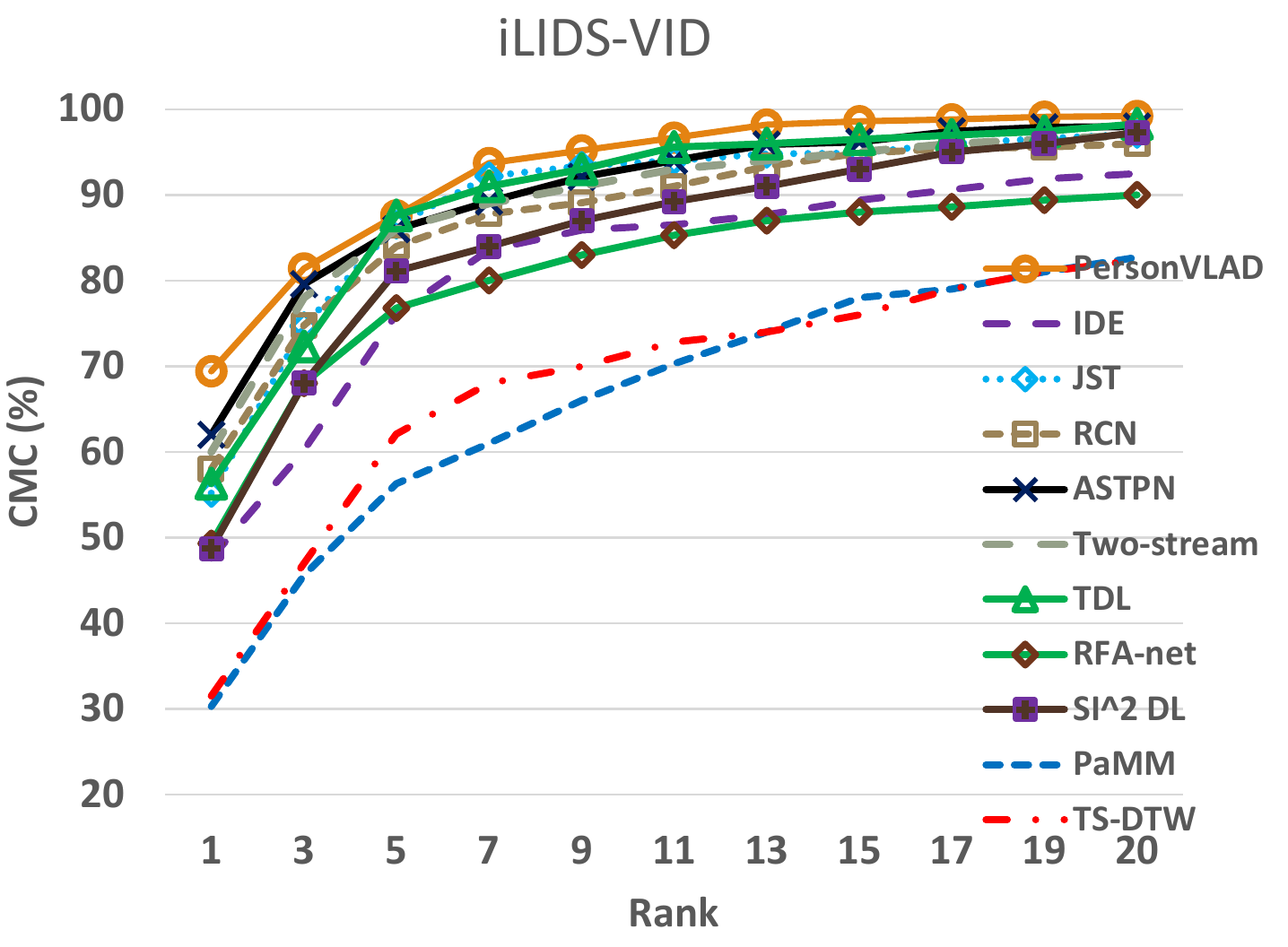} &
\includegraphics[height=4cm]{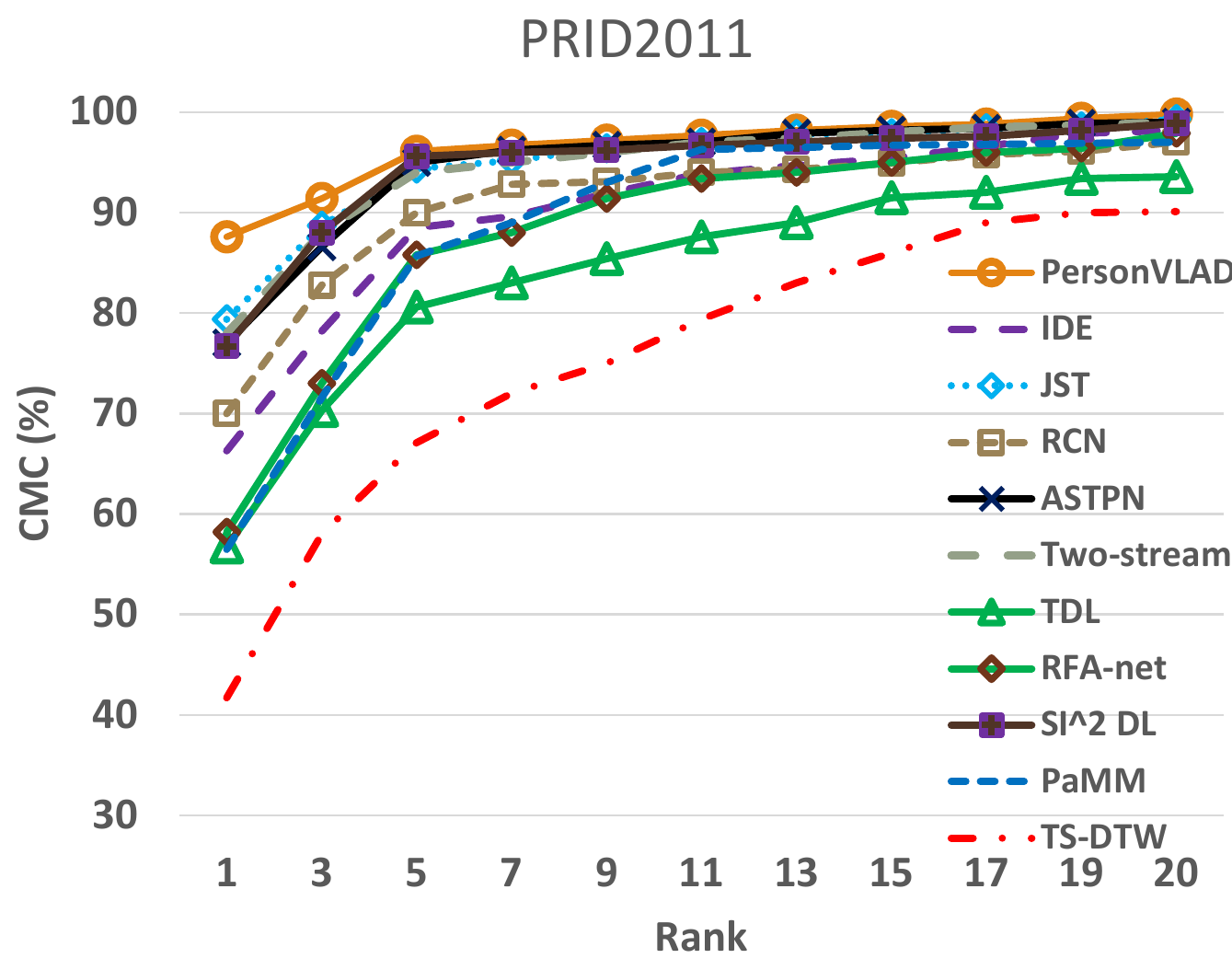} &
\includegraphics[height=4cm]{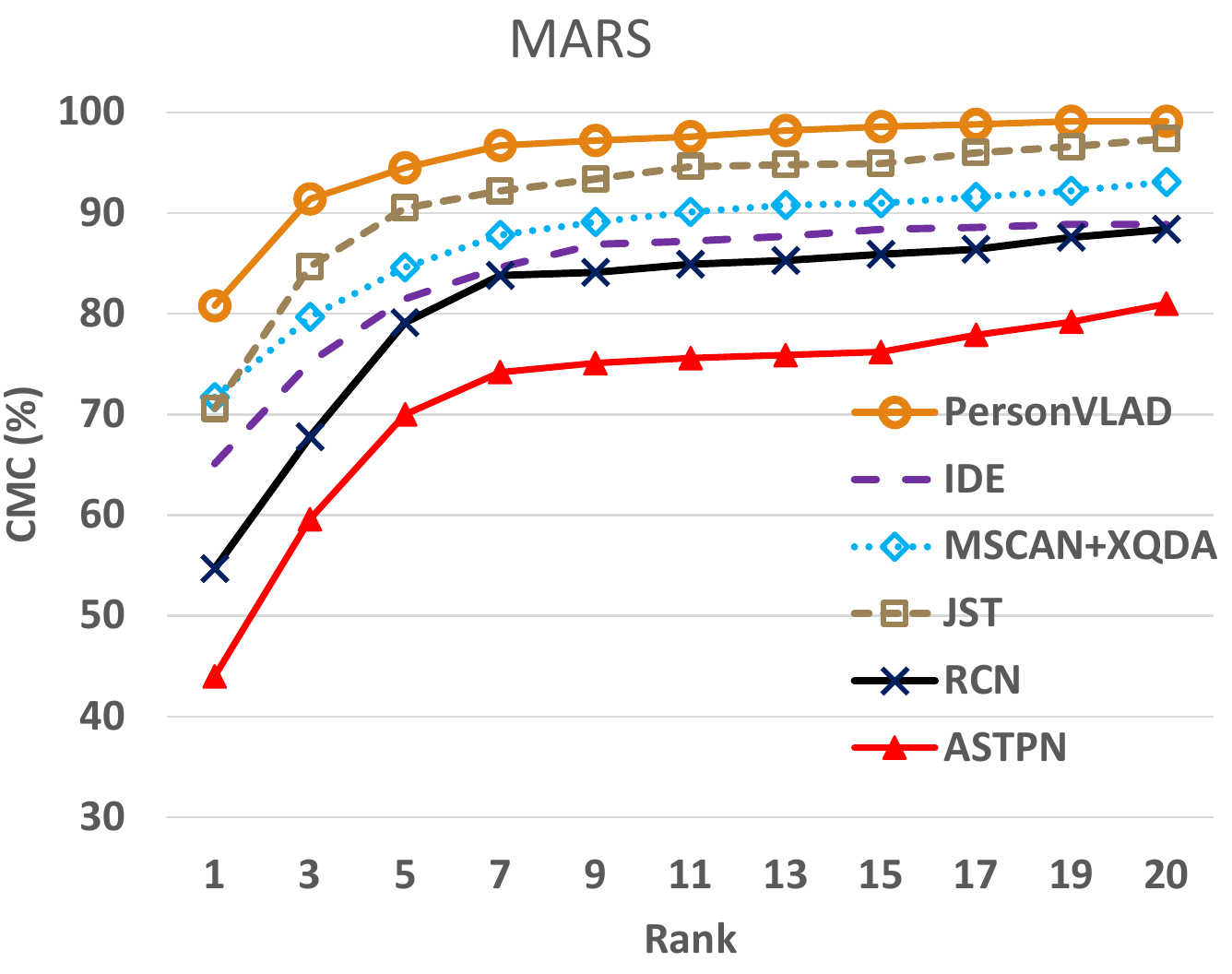}\\
 (a) iLIDS-VID & (b) PRID2011 & (c) MARS
\end{tabular}
\caption{CMC curve on the three data sets.}\label{fig:cmc-curves}
\end{figure*}

\begin{table*}[t]
\caption{Comparison with state-of-the-art methods. The results are computed in single query setting.}\label{tab:com-state}
\centering
\begin{tabular}{l|c|c|c|c|c|c|c|c|c|c|c|c}
\hline\hline
Dataset & \multicolumn{4}{c|}{iLIDS-VID} & \multicolumn{4}{c|}{PRID2011} & \multicolumn{4}{c}{MARS}\\
\cline{1-13}
Rank @ R & R = 1 & R = 5 & R = 10 & R = 20 & R = 1 & R = 5 & R = 10 & R = 20  & R = 1 & R = 5 & R = 20 & mAP \\
\hline
    HOG3D+DVR \cite{VideoRanking} & 23.3 & 42.4 & 55.3 & 68.4 & 28.9& 55.3 & 65.5 & 82.8 & 12.4 &33.2 & 71.8 & 9.6\\
    STFV3D \cite{VideoPerson} & 44.3 & 71.7 & 83.7 & 91.7 & 64.1 & 87.3 & 89.9 & 92.0 & 22.0 & 33.4 &  59.0 & 12.3\\
    TDL \cite{Top-push} & 56.3 &87.6 & 95.6 & 98.3 & 56.7 & 80.0 & 87.6 & 93.6 &-&-&-&-\\
    RFA-net \cite{RFA-net} & 49.3 & 76.8 & 85.3 & 90.0 & 58.2 & 85.8 & 93.4 & 97.9 &-&-&-&-\\
    $SI^2DL$ \cite{Video-person-ijcai16} & 48.7& 81.1 & 89.2 & 97.3 & 76.7 & 95.6 & 96.7 & 98.9 &-&-&-&-\\
    PaMM \cite{PaMM} & 30.3 & 56.3 & 70.3 & 82.7 & 56.5 & 85.7& 96.3 & 97.0 &-&-&-&-\\
    TS-DTW \cite{Video-person-matching} & 31.5 & 62.1 & 72.8 & 82.4 & 41.7 & 67.1& 79.4 & 90.1 &-&-&-&-\\
\hline
    JST \cite{Joint-spatial-temporal} & 55.2 & 86.5 & 91.0 & 97.0 & 79.4 &94.4 & 97.0 & 99.3 & 70.6& 90.0& 97.6& 50.7\\
    RCN \cite{RCNRe-id} & 58.0 & 84.0 & 91.0 & 96.0 & 70.0 & 90.0 & 95.0 & 97.0 &54.7&79.1&  88.4 & 37.4\\
    IDE \cite{MARS}+ KISSME \cite{KISSME} &47.6  & 76.1 & 86.1& 92.5 &66.3 & 88.5 & 93.9 & 98.2 & 65.0 & 81.1 &  88.9 & 45.6\\
    MSCAN \cite{Context-body}+ XQDA \cite{LOMOMetric} & - &-  &- & - & - & - & - & - & 71.7 & 86.5 & 93.1 & 56.0\\
    ASTPN \cite{ASTPN} & 62.0 & 86.0 &94.0 & 98.0 & 77.0 & 95.0 & 99.0 & 99.0 & 44.0 &70.0 & 81.0 & - \\
    Two-stream \cite{Two-stream-re-id} & 60.0 & 86.0 & 93.0& 97.0 & 78.0 & 94.0 & 97.0 & 99.0 & - &- & - &- \\
  \hline
  PersonVLAD & \color{red}$\mathbf{69.4}$ & 87.6 & \color{red}$\mathbf{96.7}$ & \color{red}$\mathbf{99.2}$ & \color{red}$\mathbf{87.6}$ & \color{red}$\mathbf{96.1}$ & \color{red}$\mathbf{98.7}$ & \color{red}$\mathbf{99.8}$ & \color{red}$\mathbf{80.8}$ & \color{red}$\mathbf{94.5}$ & \color{red}$\mathbf{99.0}$ & \color{red}$\mathbf{63.4}$\\
  PersonVLAD + XQDA \cite{LOMOMetric} & \color{red}$\mathbf{70.7}$ & \color{red}$\mathbf{88.2}$ & \color{red}$\mathbf{97.1}$ & \color{red}$\mathbf{99.2}$ & \color{red}$\mathbf{88.0}$ & \color{red}$\mathbf{96.2}$ & 98.6 & 99.7 & \color{red}$\mathbf{82.8}$ & \color{red}$\mathbf{94.9}$ & \color{red}$\mathbf{99.0}$ & \color{red}$\mathbf{64.7}$ \\
\hline
\end{tabular}
\end{table*}

\subsection{Comparison with State-of-the-arts Methods}

In this experiment, we compare the proposed 3D PersonVLAD with state-of-the-arts in video based person re-ID. The experimental results are summarized in Table \ref{tab:com-state} and Fig.\ref{fig:cmc-curves}.

\textbf{MARS}: In MARS, several state-of-the-art methods are compared, including JST \cite{Joint-spatial-temporal}, RCN \cite{RCNRe-id}, IDE \cite{MARS}, MSCAN \cite{Context-body}, and ASTPN \cite{ASTPN}. Compared with the methods based on the combination of CNNs and RNNs, such as RCN \cite{RCNRe-id}, IDE \cite{MARS} and ASTPN \cite{ASTPN}, the proposed approach can preserve the spatiotemporal information through all layers of the networks to avoid the collapsing of temporal priors by using 2D ConvNets. Moreover, these current methods are limited in shorter sequences and they even rely on a pre-selection step to select informative frames to conduct the feature learning. In contrast, our PersonVLAD is able to deal with the entire video to produce a global yet highly discriminative representation. Compared with full body representations, such as JST \cite{Joint-spatial-temporal}, our method aims to aggregate body part based representations to combat the misalignment issue. MSCAN \cite{Context-body} considers the learning of features over both full body and body parts, however, they use the spatial transformation networks to encode spatial constraints and introduce additional parameters.

\textbf{iLIDS-VID and PRID2011}: For the two small data sets, we fine-tune the pre-trained model on MARS on their respective training split and then compute the CMC values on testing. We can observe that our approach is able to achieve state-of-the-art results, e.g., rank-1=69.4 vs rank-1=62.0 of ASTPN \cite{ASTPN} on iLIDS-VID and rank-1=87.6 vs rank-1=79.4 of JST \cite{Joint-spatial-temporal} on PRID2011. We suspect the main reason accountable for the marginal improvement on PRID2011 is the scarcity of training samples to optimize the networks. As for iLIDS-VID, our method can achieve the highest accuracy of rank-1=69.4, outperforms the state-of-the-art ASTPN \cite{ASTPN} by a margin of 7.4\%, and the learned global features can improve the recognition rate further if they are combined with a metric learning algorithm, e.g., XQDA \cite{LOMOMetric}.

\subsection{Computational Analysis and Scalability Study}

\begin{figure}[t]
\centering
\begin{tabular}{c}
\includegraphics[height=4cm]{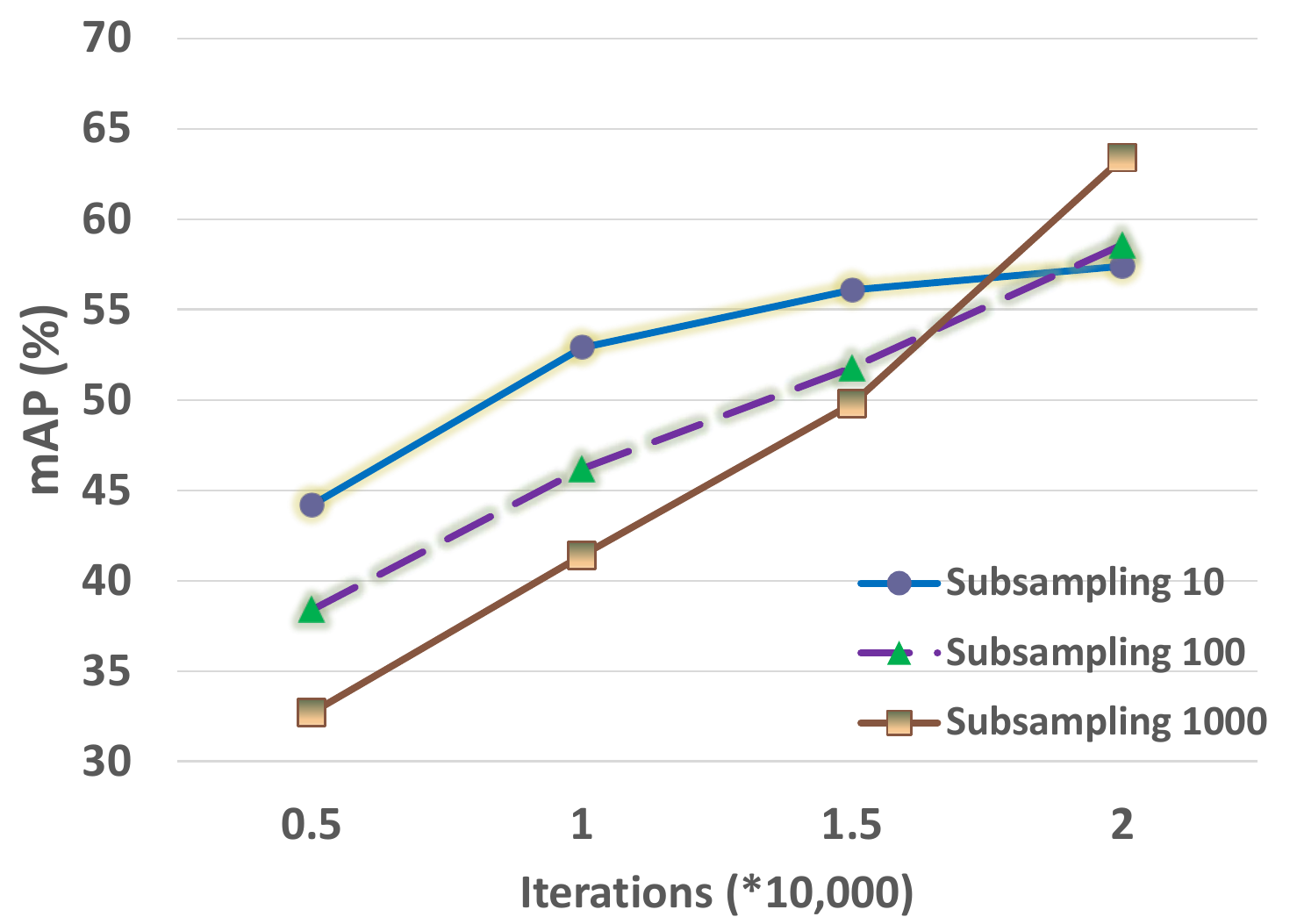} \\
\end{tabular}
\caption{Scalability: The study on the sub-sampling effect on the loss function optimization.}\label{fig:sub-sample}
\end{figure}

\begin{table}[t]\small
  \centering
  \caption{The comparison of computational cost on MARS.}\label{tab:running-time}
  {
  \begin{tabular}{l|c|c|c}
  \hline
    Method  &  Training & inference  & mAP \\
  \hline
  PersonVLAD & 1.2 days & 1182.4fps & 63.4\\
  C3D \cite{3DConvNets} & 3.4 days & 313.9fps & 58.9\\
  2D ConvNets \cite{Two-stream-fusion} & - & 1.2fps & 51.4\\
  \hline
  \end{tabular}
  }
\end{table}

In this experiment, we compare the computational cost with C3D \cite{3DConvNets}, and a temporal stream network based on 2D ConvNets \cite{Two-stream-fusion} in terms of running time in training and inference. The comparison results on the MARS dataset are reported in Table \ref{tab:running-time}. In inference, the PersonVLAD runs at 1182.4fps on a Titan X GPU and 16 CPU cores, which is 3.8 faster than C3D and efficiently applicable in real-time analysis. This is mainly because the aggregation layer with 3D body part detection is amount to reducing the spatial resolution, and thus the learned video feature is still computationally efficient with increasing the temporal extent. For the 2D ConvNets \cite{Two-stream-fusion}, the method uses Brox's optical flows \cite{Brox} as temporal inputs that are computed by stacking every 10 frames. The fusion on spatial dimensions is implemented by a convolutional fusion which can perform best as opposed to sum fusion and concatenation \cite{Two-stream-fusion}, however, it is running at a much longer training time with a large number of parameters ($\sharp$97.58M \cite{Two-stream-fusion}).

To further study the scalability of network in training which is the computational concern caused by the time-consuming computation on the partition function in Eq. \eqref{eq:loss}, we employ the sub-sampling strategy on the labelled and unlabelled identities as suggested by the Xiao \etal \cite{OIM}. This validation is performed by training the network on MARS with sub-sampling size of 10, 100, and 1000. The mAP curves are shown in Fig. \ref{fig:sub-sample}. It can be observed that sub-sampling leads to faster convergence rate even though some slight drop on performance. It indicates that our network is scalable to large-scale dataset and video features can be efficiently produced.